# TrafficSafetyGPT: Tuning a Pre-trained Large Language Model to a Domain-Specific Expert in Transportation Safety

Ou Zheng, PhD (Corresponding Author)
Department of Civil, Environmental & Construction Engineering University of Central Florida, Orlando, FL, 32816, USA
Email: ou.zheng@ucf.edu

Mohamed Abdel-Aty, PhD
Department of Civil, Environmental & Construction Engineering University of Central Florida, Orlando, FL, 32816, USA
Email: m.aty@ucf.edu

Dongdong Wang , PhD
Department of Civil, Environmental & Construction Engineering University of Central Florida, Orlando, FL, 32816, USA
Email: Dongdong.Wang@ucf.edu

Chenzhu Wang , PhD
Department of Civil, Environmental & Construction Engineering University of Central Florida, Orlando, FL, 32816, USA
Email: chenchen9989@sohu.com

Shengxuan Ding
Department of Civil, Environmental & Construction Engineering University of Central Florida, Orlando, FL, 32816, USA
Email: Shengxuan.Ding@ucf.edu

**Abstract**

Large Language Models (LLMs) have shown remarkable effectiveness in various general-domain natural language processing (NLP) tasks. However, their performance in transportation safety domain tasks has been suboptimal, primarily attributed to the requirement for specialized transportation safety expertise in generating accurate responses [1]. To address this challenge, we introduce TrafficSafetyGPT, a novel LLaMA-based model, which has undergone supervised fine-tuning using TrafficSafety-2K dataset which has human labels from government produced guiding books and ChatGPT-generated instruction-output pairs. Our proposed TrafficSafetyGPT model and TrafficSafety-2K train dataset are accessible at
https://github.com/ozheng1993/TrafficSafetyGPT

## 1. Introduction

In the realm of natural language processing (NLP) and large language models, a surge in advancements has unfolded a plethora of potential applications. This rapid development, spearheaded by pre-trained large language models like OpenAI's ChatGPT and its derivatives, has drastically augmented our capabilities in language comprehension, generation, and interactivity. The foundational strength of these models lies in their pre-training on extensive and diverse datasets, empowering them to decipher intricate language patterns and contextual interconnections.

Nevertheless, while these pre-trained models exhibit commendable proficiency across an array of tasks, their generic nature could constrain their efficacy in niche applications, such as transportation safety. The training of a Large Language Model (LLM) from scratch necessitates the ingestion of vast, heterogeneous data to master sophisticated language patterns and relationships, which in turn requires potent hardware like high-performance GPUs or TPUs to manage the colossal computational load during training. Furthermore, the training duration may extend from a few days to weeks, subject to the model's size and complexity.

This research paper aims to address this conundrum by concentrating on the fine-tuning of a pre-trained offline large language model, Meta AI's LLaMA, to cultivate a subject-matter expert in transportation safety. The process capitalizes on the pre-trained model's inherent knowledge and linguistic proficiency, amalgamating it with a comprehensive transportation safety dataset gleaned from government-produced guidelines and instruction-output pairs generated by ChatGPT. The objective is to engineer a specialized language model capable of producing precise, context-sensitive, and safety-conscious responses in traffic-related situations. The novelty of TrafficSafetyGPT is its potential to redefine how language models augment traffic safety and the broader transportation field. Instead of constructing a model from the ground up, our methodology refines the existing linguistic aptitude through fine-tuning with small dataset TrafficSafety-2k, yielding a more efficient allocation of computational resources and training time. Moreover, this tailored model assures that the output aligns with ratified traffic safety standards and guidelines, thus ensuring its practical applicability.

## 2. Related Works

### 2.1 Large Language Models

Large Language modeling (LLM) offers impressive capabilities in natural language processing and artificial intelligence. There were many models which have been proposed in the current analysis, including GPT4 [2], Palm2 [3] , Claude [4], Cohere [5], Falcon [6], and Meta LLaMA [7].Some of the key roles of these large language models include natural language understanding, language generation, reasoning tasks, text processing, language tutoring and learning. The introduction of the GPT-4 model marked a significant advancement in natural language understanding and generation [8]. This latest iteration showcased remarkable abilities across various domains, including complex reasoning, comprehensive comprehension, advanced coding capabilities, proficiency in diverse academic exams, and even human-level performance in numerous tasks [9]. Its impressive attributes have garnered widespread attention and acclaim. For reasoning tasks, PaLm2 was an advanced language model from Google with enhanced multilingual capabilities, improved reasoning skills, and proficient coding capabilities, having been trained on a diverse dataset encompassing over 100 languages, scientific papers, web pages containing mathematical expressions, and publicly available source code datasets [3]. In the aspect of text processing, Claude offers a wide range of use cases, such as summarization, search, creative and collaborative writing, Q&A, coding, and much more. Just like ChatGPT, Claude boasts a user-friendly chat interface and API, which developers can access through our developer console. Its outstanding abilities in handling various conversational and text processing tasks are complemented by its exceptional reliability and predictability [10]. LLM impressed many users by virtue of its exceptionally superior performance based on billions of parameters, which provides a solid foundation for the fine-tuning of models. For example, Cohere offers offer multiple models, varying in size from small to large, with as few as 6 billion parameters to the larger models trained on an extensive 52 billion parameters [11]. Moreover, Falcon is open-source and released under the Apache 2.0 license [12]. This permits users to utilize the model for commercial purposes without any royalties or restrictive limitations. In addition, the official release of LLaMA models comes in different sizes, ranging from 7 billion to 65 billion parameters. Numerous developers are utilizing LLaMA for fine-tuning and creating exceptional open-source models [13]. However, it is important to note that LLaMA has been made available for research purposes exclusively. As an extension of LLaMA, Meta LLaMA 2 introduces pertained and fine-tuned LLMs with 7B, 13B, and 70B parameters. These models improve upon LLaMA 1 with 40% more tokens, a context length of 4,000 tokens for better language comprehension, and grouped-query attention for faster inference in the 70B model [14]. Given this, the existing Large Language Models were developed to be more academic and human-level in many aspects of studies. However, in the field of traffic safety, a limited body of research efforts has been proposed to analyze trajectory prediction [15], safety planning and design [16], safety rules [17], etc. And LLaMA is proven to be effective and available in research analysis, which will be used as the base model of the TrafficSafetyGPT in the current study.

**2.2 Pre-trained Models with Domain Knowledge Fine-tuning**

Due to the swift advancements in LLM and the increasing demands for practical applications across diverse fields, researchers recognized the need to enhance the performance of LLMs by fune turning the models based on domain knowledge. For example, researchers have acknowledged the imperative to improve the performance of LLMs in medical applications, where precision is of utmost importance, primarily due to their limited domain-specific knowledge. To address this issue, they developed ChatDoctor, a specialized model tailored specifically for the biomedical field. The main objective of ChatDoctor is to enhance the understanding of LLMs in this domain. This endeavor underscores the significance of integrating domain-specific knowledge to attain superior outcomes in medical applications [18].
    Another notable study focused on enhancing LLMs' performance in the specific domain through domain-specific knowledge transfer with different languages. As an illustration, researchers effectively fine-tuned the LLaMA model with Chinese medical knowledge, which contributed to improved comprehension and performance in Chinese LLM applications [19]. This research emphasized the importance of domain-specific tuning to optimize LLMs for specific fields [20].
Several technical approaches were explored to enhance domain knowledge transfer and fine-tuning of LLMs. Effective techniques such as soft fine-tuning [21], large-scale fine-grained categorization [21], and domain-specific transfer learning were employed to optimize LLM performance in specialized domains(). Additionally, the LLaMA-adapter method demonstrated efficient fine-tuning of language models with zero-init attention, contributing to better adaptability in domain-specific tasks [22]. There are some domain knowledge datasets that support LLM fine-tuning, such as Alpaca, which contributes 52k instruction-following data points by leveraging Self-Instruct techniques to encode specific instructions within conversations [23] .The HealthCareMagic-100k dataset encompasses 100k real-world patient-physician conversations, providing valuable insights into authentic medical interactions [18]. On the other hand, GenMedGPT-5k enriches dataset diversity with 5k synthetic conversations between patients and physicians [18].

## 3 TrafficSafetyGPT

### 3.1 Foundation Model

We selected a lightweight version of ChatGPT– LLaMA as a foundation model. A family of models derived from LLaMA represents a cutting-edge collection of open-access large language models ranging from 7B to 70B parameters (7B, 13B, 70B) introduced by Meta AI, showcasing significant advancements in natural language processing research. However, LLaMA is made accessible under a non-commercial license, restricting its usage exclusively to academics with specific credentials. However, recent developments have been unveiled by Meta, with the announcement of LLaMA 2—a new family of AI language models that also falls under the source-available category but distinguishes itself with the inclusion of a commercial license, enabling broader usage beyond non-commercial purposes.

### 3.2 Transportation Safety Knowledge

Government departments and agencies produce guiding books pertaining to specific domains, such as healthcare, education, finance, and more. This allows Large Language Models (LLMs) to grasp domain-specific language and terminologies, enhancing their ability to communicate effectively in specialized contexts. Moreover, these government guiding books encompass authoritative and accurate information on various subjects, including laws, regulations, policies, and procedures. By training LLMs on this data, they gain reliable knowledge, enabling them to deliver well-informed and precise responses. Additionally, exposure to real-world contexts and scenarios in these guiding books deepens LLMs' understanding of practical language applications, leading to more contextually relevant and useful outputs. Furthermore, the information contained in these books often involves decision-making processes, problem-solving guidelines, and strategies for complex situations. Consequently, LLMs trained on such data can offer valuable insights and recommendations in diverse governance and administrative tasks. We have manually labeled 665 rows of data from the perspective of traffic management, traffic operation, and drivers. The ground truth follows the government book as fine-tuning instruction data incorporating the NSTHA Model Minimum Uniform Crash Criteria (MMUCC) Guideline Fourth edition and FHWA's Highway Safety Manual (HSM) in the training process of LLMs contributes to creating safer, more informed, and reliable language models in the domain of traffic safety and transportation.

- NSTHA Model Minimum Uniform Crash Criteria (MMUCC) Guideline Fourth edition MMUCC was established with the aim of fostering increased uniformity and consistency. Its purpose is to offer State and Local agencies a standardized set of data variables related to motor vehicle traffic crashes, encouraging them to consider these elements for collection.

The key components of MMUCC Guidelines cover various aspects of crash data collection and reporting, including 1689 rows of data. The first type is the definition. The guidelines provide standardized definitions for each data element, ensuring that different agencies interpret and record the information in the same way. In addition, MMUCC outlines the classification of crashes based on severity, collision type, and other relevant factors to enable meaningful analysis and comparison. Lastly, the guidelines offer recommendations on reporting formats and methods, encouraging efficient data exchange and sharing among agencies.

- FHWA The Highway Safety Manual (HSM)

  The Highway Safety Manual (HSM), published by the American Association of State Highway Transportation Officials (AASHTO), is a valuable resource for evaluating traffic safety on existing or proposed roadways. It offers a science-based and technical approach that helps state and local agencies analyze safety without relying on guesswork. The HSM promotes a safety management process that involves identifying high-crash locations, analyzing contributing factors, and implementing countermeasures to mitigate risks and improve overall safety. 311 rows of data related to the safety management process are labeled in our work. It utilizes a data-driven approach and employs decision-making strategies to allocate resources more efficiently by focusing on locations with the greatest safety improvement potential.

**Table 1 TrafficSafety-2K number of data**

|  | NSTHA Model Minimum Uniform Crash Criteria (MMUCC) Guideline Fourth edition | FHWA The Highway Safety Manual (HSM) | ChatGPT generated Data |
|---|---|---|---|
| Number of label | 1689 | 311 | 2000 |

However, books have a limited amount of labeled data, and relying solely on them for training a domain-specific language model may not be sufficient. rows of data related to transportation safety were generated by ChatGPT. Using generated conversations from ChatGPT with content from government departments and agencies guiding books as training data for a pre-trained large language model to become a domain-specific expert in transportation safety offers several benefits. The specialized nature of generated conversations ensures the model focuses solely on transportation safety topics while incorporating the comprehensive information from guiding books helps it gain an authoritative understanding of the subject. By simulating real-world scenarios, the model learns to respond appropriately to various safety concerns, enhancing its practical applicability. Moreover, using standardized information from government sources promotes consistency and accuracy in the model's responses while mitigating potential biases.

Achieving optimal performance necessitates a balanced approach, combining both generated and real data, and fine-tuning the model to ensure it provides reliable and knowledgeable insights in the transportation safety domain. We name this dataset TrafficSafety-2K.

### 3.3 Knowledge-based Instruction Data

**Table 2 Example of Knowledge-based Instruction Data**

| Type | Input | Knowledge |
|---|---|---|
| Definition | What is the definition of a van in Motor Vehicle Traffic Crashes? | A van is a motor vehicle consisting primarily of a transport device that has a gross vehicle weight rating (GVWR) of 10,000 pounds or less and is basically a "box on wheels" that is identifiable by its enclosed passenger and/or cargo area, step-up floor, and relatively short (or nonexistent) hood. Vans are classified by size based on frame type and overall vehicle body width. Before classification, vehicle width should be rounded to the nearest inch. |
| Inclusions | What are the inclusions of Trafficway in Motor Vehicle Traffic Crashes? | Within areas with guarded entrances, such as military posts or private residential developments, land ways are trafficways if the guards customarily admit public traffic. Privately constructed and/or maintained road open to the public for moving persons or property for transportation purposes. Local road in a residential development, which is open to the public. Land way providing vehicular access and/or circulation from a trafficway to a business open to the public |
| Exclusions | What are the exclusions of a commercial motor vehicle in Motor Vehicle Traffic Crashes? | Privately owned motor vehicle providing private transportation of personal property or people. |
| Categories | What is the guide to the classification of Persons by injury severity in Motor Vehicle Traffic Crashes? | The injury classification applies to any person involved in road vehicle crashes while either in or out of a road vehicle. The categories are so defined that, for the most part, neither medical attention nor special tests are required for classification. Classification usually can be done by ordinary observation at the time of the crash or from information submitted on the crash report. |

| Examples | What are the Examples of transit bus in Motor Vehicle Traffic Crashes? | City metro or ride-on bus.Trolley (on highway tires). |
|---|---|---|
| Guidance | How do you deal with collisions and emergencies? | Personal safety and the safety of any passengers should always be your first consideration. Use your hazard warning lights and high-visibility clothing to make sure you and your vehicle can be seen by other road users. Collisions that result in injury while driving for work should be reported to the Gardai, as appropriate, and your line manager immediately. Details should be recorded on a preliminary incident report form. |

## 4 Experiment

We select ChatGPT as a baseline foundation model for text generation assessment. For efficient implementation, a pre-trained LLaMA-7B foundation model is chosen as a practical alternative and fine-tuned with domain knowledge. To improve domain-specific knowledge with limited data, we only tuned the last two layers. This fine-tuning strategy is developed for the reasons as follows. First, although our traffic-safety-domain specific data are adequate, the number is not comparable to the scale of a large language knowledge database, like Wikipedia. Fine-tuning all layers with limited data can easily result in overfitting. Second, the original features in the pre-trained LLaMA are still helpful for language generation, like reasoning and fluency. Third, this fine-tuning on the last two layers saves lots of both space and time cost, which enables efficient training and practical implementation.

Based upon the designed strategy , we fine-tuned the LLaMA-7B model on the Lambda server with 8 RTX TITAN GPUs, 128 CPU cores, and 500GB RAM. The fine-tuning process is completed with three training epochs which costs ~70 CPU hours. The training is implemented with the hyper parameters as follows: the batch size of 16, the learning rate of $2 \times 10^{-5}$, the epoch of 3, the maximum sequence length of 152 tokens, and the warm-up ratio of 0.03 without weight decay.

### 4.1 Base model

LLaMA is one of important foundational large language models created by Meta AI. Just like its counterparts in the field, LLaMA operates on the principle of accepting a word sequence as input and employing its predictive capabilities to generate subsequent words recursively, ultimately producing coherent and contextually appropriate text.

In the study, we follow the Stanford instruction-following LLaMA Model train processing: Alpaca [24]. Figure 1 demonstrates the process for instruction-following, employing a methodology that builds upon the self-instruct approach utilized in MUCC and HSM. To initiate this process, we utilized a set of 1k instruction-output pairs that were previously human-labeled in those specific books. Subsequently, we leveraged the In ChatGPT model, prompting it to generate additional instructions while utilizing the aforementioned human-labeled dataset as in-context examples. Figure 2 shows the example of human labels and ChatGPT generate instruction-output pairs.

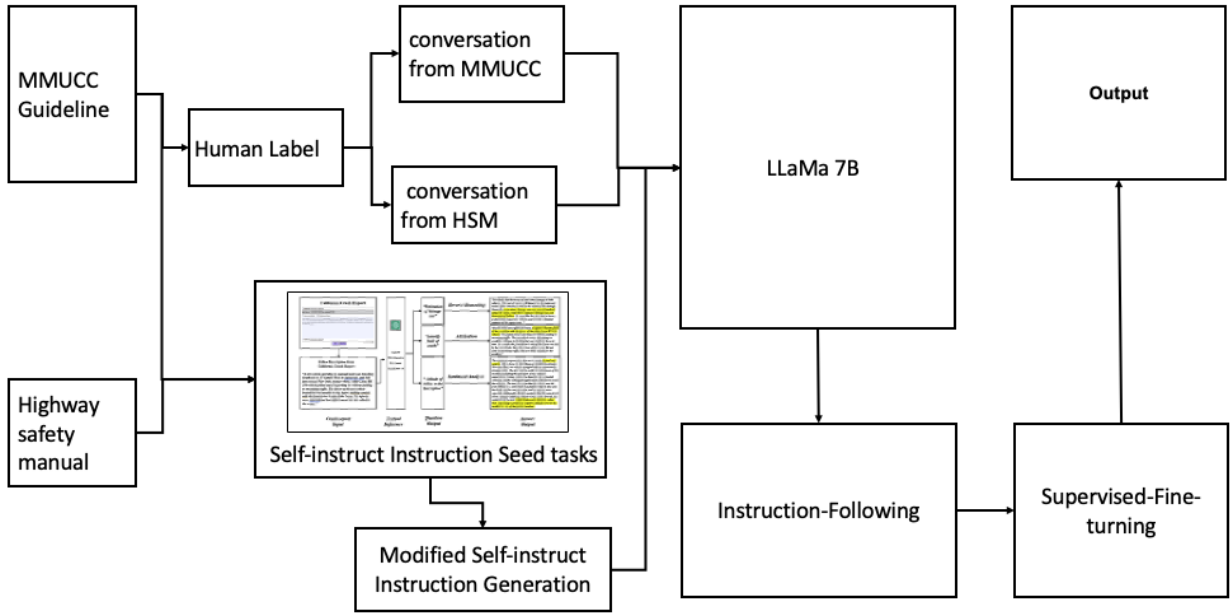

Figure 1 Data generation process for instruction-following.

| Example of Human Label of MMUCC | Example of Human Label of HSM | Example of ChatGPT generate of MMUCC | Example of ChatGPT generate of HSM |
|---|---|---|---|
| **Instruction**: You are a police officer at the crash Scene<br><br>**Input**: What is the definition of property in Motor Vehicle Traffic Crashes<br><br>**output**: Property is any physical object other than a person. | **Instruction**: You are a traffic engineer work for DOT<br><br>**Input**: What is weather conditions and seasonal hazards?<br>**output**: The weather can cause all sorts of driving hazards, especially reduced visibility and slippery road surfaces. You need to adjust your driving accordingly. Listen to local/national weather broadcasts and travel bulletins. | **Instruction**: You are a police officer at the crash Scene<br><br>**Input**: Is person consider as property in traffic crashes?<br><br>**output**: No, Property should be physical objects | **Instruction**: You are a traffic engineer work for DOT<br><br>**Input**: What can cause all sorts of driving hazards<br><br>**output**: The weather can cause all sorts of driving hazards, |

Figure 2 Example of human label and ChatGPT generate instruction-following.

## 4.2 Metrics

We evaluate the proficiency of generated texts in traffic safety with domain questions of different kinds, including definition, inclusion, exclusion, category, example, and guidance. The generated answers to these questions are evaluated by several popular and reliable quantitative metrics, which examines the results from coverage, fluency, and brevity. For coverage, we selected BLEU, ROUGE, and BERTScore. For fluency, we chose BLEURT. For brevity, we count the words of the generated sentences.

**Coverage.** BLEU (bilingual evaluation understudy) is a popular metric for natural language generation. This algorithm is first proposed to evaluate machine translation performance. Since computer generated texts can be taken as a different language system from human language, this score can also reflect the quality of machine-generated texts. BLEU ranges from 0 to 1, which indicates the similarity between human languages and machine generated texts. However, this score is quantified heavily based upon word matching on golden labels, which can yield some bias. For example, a correctly generated machine language can still show a low BLEU score because of the paraphrase; meanwhile, just one word change may yield totally different meaning, but the BLEU score can stay similar.

ROUGE (Recall-Oriented Understudy for Gisting Evaluation) is an algorithm to assess text summarization and machine translation. The metric can evaluate the information coverage of automatically produced texts on a reference passage which is usually a human-written summary. ROUGE relies on n-gram modeling which considers word overlapping between generated and reference texts. The common ROUGE scores include ROUGE-1 (overlap of unigrams), ROUGE-2 (overlap of bigrams), and ROUGE-L (longest common subsequence based statistics). Since ROUGE is proposed to solve a summarization problem, it is very effective and reliable to assess the information coverage. However, it can also result in the bias due to lack of semantic meaning understanding. Therefore, like BLEU, a high ROUGE score may lead to a sentence with a different meaning.

BERTScore is a recently proposed metric based upon a milestone large language model of BERT. It relies on the prediction results from BERT to evaluate candidate sentences and reference sentences by tokens. BERTScore computes the token similarity with the context embeddings, which replaces word exact match. This evaluation is more focused on comparison on a vectorized embedding space instead of a common letter system, which is more suitable for large language model evaluation. Since BERT is developed upon mask modeling, its word embeddings can reflect mask filling accuracy and help effectively assess information coverage. However, one of its important weaknesses is language model dependency, where model bias will be induced in the evaluation process.

**Fluency.** BLEURT (Bilingual Evaluation Understudy with Representations from Transformer) is adopted to evaluate the fluency of generated texts. This evaluation metric is developed based upon BERT, which is derived from the outputs of a robust BERT model pre-trained by perturbed synthetic sentences from Wikipedia. This metric can reflect the paraphrase level and language fluency. Although it relies on a large language model and may induce model bias, its fluency assessment is more reliable due to pretraining on a large set of synthetic data.

**Brevity.** We adopt the Word Count to assess the brevity of the expression, which is straightforward and also popular in summarization evaluation. Fewer words indicate better brevity of the generated texts.

### 4.3 Results

To evaluate the efficacy of TrafficSafetyGPT, we conduct extensive experiments on various transportation safety-related tasks. We compare its performance with the state-of-the-art non-specialized LLM of LLaMA. The results justify that TrafficSafetyGPT consistently outperforms on different specific tasks. This observation results from the success of domain-specific fine-tuning, especially with more structural presentation and professional expression. It is obtained that the information coverage of TrafficSafetyGPT is significantly higher than LLaMA, even on the BLEU score. The performance also shows that the enumeration tasks, like inclusions, exclusions, categories, and examples, are more challenging for language modeling because these tasks are sensitive to exact match with each single word. The performance gain of TrafficSafetyGPT on these tasks results from a larger domain specific terminology database, which helps enhance including more professional terms. For guidance, LLaMA exhibits wordy expressions with higher word count since the generalized domain does not provide technical information focus. TrafficSafetyGPT yields briefer expressions after fine-tuning on domain-specific contexts.

**TABLE 3 The examples of TrafficSafetyGPT generated texts.**

| Instruction | Input | Response |
|---|---|---|
| If you are a transportation engineer, please answer the transportation questions based on the student's description. | What is the consumer vehicle? | A Nissan Leaf. A Chevy Spark, Chargé, or Volt. An Underwood Navigator. A Smart ForTwo. A Rhino. A parking lot Fuller hog. A metallic or all-plastic bus vault. A custom made on a truck chassis. An exciting electric or hybrid road vehicle with speed postings or numbers. |
| If you are a transportation engineer, please answer the transportation questions based on the student's description. | What is the working vehicle? | A working vehicle is any vehicle designed primarily for carrying property. |
| If you are a transportation engineer, please answer the transportation questions based on the student's description. | What is the difference between a consumer vehicle and a working vehicle? | A consumer vehicle is any vehicle other than a working vehicle. A consumer vehicle is usually stored in its owner's home or garage most of the time. |
| If you are a transportation engineer, please answer the transportation questions based on the student's description. | Is patrol a working vehicle? | No, a patrol is a transport vehicle. |

**Table 4 Comparison of TrafficSafetyGPT output**

| Type of task | ChatGPT | TrafficSafetyGPT |
|---|---|---|
| Definitions | | |
| BLEU | 7.38 | **36.04** |
| ROUGE-1 | 40.54 | **80.66** |
| ROUGE-2 | 10.11 | **50.99** |
| ROUGE-L | 39.10 | **78.98** |
| BERT-P | 84.74 | **96.75** |
| BERT-R | 91.55 | **97.98** |
| BLEURT | 47.34 | **69.30** |
| Word Count | 74 | **37** |
| Inclusions | | |
| BLEU | 2.39 | **32.75** |
| ROUGE-1 | 38.33 | **81.47** |
| ROUGE-2 | 15.28 | **67.76** |
| ROUGE-L | 37.95 | **81.09** |
| BERT-P | 96.35 | **98.07** |
| BERT-R | 96.45 | **98.00** |
| BLEURT | 35.54 | **58.55** |
| Word Count | 81 | **42** |
| Exclusions | | |
| BLEU | 4.56 | **31.64** |
| ROUGE-1 | 29.93 | **78.31** |
| ROUGE-2 | 13.14 | **60.20** |
| ROUGE-L | 28.77 | **77.99** |
| BERT-P | 96.79 | **98.11** |
| BERT-R | 96.68 | **98.09** |
| BLEURT | 26.48 | **50.15** |
| Word Count | 90 | **47** |
| Categories | | |
| BLEU | 1.79 | **30.47** |
| ROUGE-1 | 26.03 | **77.49** |
| ROUGE-2 | 11.31 | **55.21** |
| ROUGE-L | 26.02 | **77.23** |
| BERT-P | 96.85 | **98.21** |
| BERT-R | 96.76 | **98.01** |
| BLEURT | 35.09 | **59.01** |
| Word Count | 91 | **51** |
| **Examples** | | |
| BLEU | 2.03 | **30.90** |
| ROUGE-1 | 32.50 | **80.39** |
| ROUGE-2 | 13.44 | **61.11** |
| ROUGE-L | 31.99 | **80.31** |
| BERT-P | 96.77 | **98.10** |
| BERT-R | 96.07 | **97.97** |
| BLEURT | 38.97 | **62.38** |
| Word Count | 101 | **46** |
| **Guidance** | | |
| BLEU | 3.92 | **33.24** |
| ROUGE-1 | 28.39 | **75.39** |
| ROUGE-2 | 10.30 | **56.13** |
| ROUGE-L | 27.98 | **74.97** |
| BERT-P | 96.77 | **98.19** |
| BERT-R | 96.47 | **98.03** |
| BLEURT | 40.44 | **64.73** |
| Word Count | 123 | **50** |

## 5. CONCLUSIONS

In this paper, we propose the TrafficSafetyGPT, a LLaMA-based model that addresses the limitations faced by Large Language Models in transportation safety domain tasks. By employing supervised fine-tuning with the TrafficSafety-2k dataset, human labels from the government produce guiding books and ChatGPT-generated instruction-output pairs. TrafficSafetyGPT demonstrates superior performance in generating responses with reliable traffic safety knowledge, allocating efficient computational resources and saving training time. Our research opens new avenues for enhancing language models' capabilities in transportation safety domains and contributes to the advancement of natural language processing applications in specialized fields.

## 6. AUTHOR CONTRIBUTIONS

The authors confirm contribution to the paper as follows: study conception and design: Ou Zheng, Mohamed Abdel-Aty,Dongdong Wang,Shengxuan Ding,Chenzhu Wang; data collection and processing: Ou Zheng, Shengxuan Ding, Chenzhu Wang; analysis and interpretation of results: Chenzhu Wang, Dongdong wang; draft manuscript preparation: Ou Zheng, Mohamed Abdel-Aty,Dongdong Wang, Shengxuan Ding,Chenzhu Wang. All authors reviewed the results and approved the final version of the manuscript.

## 7. Availability

The TrafficSafetyGPT model and the curated training dataset are publicly accessible at https://github.com/ozheng1993/TrafficSafetyGPT for research and non-commercial purposes. It is crucial to emphasize that the accuracy of responses generated by large language models cannot be assured. The safety knowledge presented by these models should not be considered a replacement for professional engineer advice.We encourage researchers and practitioners to leverage this resource to further explore and improve language models' capabilities in transportation safety domains.